\documentclass[letterpaper,twocolumn,10pt]{article}
\usepackage{usenix2019_v3}
\usepackage{custom_macros}
\usepackage{capt-of}
\usepackage{bm}
\usepackage{booktabs}
\usepackage{tabularx}
\usepackage{subfigure}
\usepackage{xcolor}
\usepackage{tikz}
\usepackage{amsmath}
\usepackage{hyperref} 
\usepackage[flushleft]{threeparttable} 
\usepackage{booktabs,caption}

\begin{document}

\date{}

\title{\Large \bf Did the Neurons Read your Book? \\
        Document-level Membership Inference for Large Language Models}

\author{
 {\rm Matthieu Meeus}\\
 Imperial College London
 \and
 {\rm Shubham Jain}\\
 Sense Street
  \and
  {\rm Marek Rei}\\
 Imperial College London
 \and
  {\rm Yves-Alexandre de Montjoye}\\
 Imperial College London
} 

\maketitle

\pagenumbering{gobble}

\begin{abstract}

With large language models (LLMs) poised to become embedded in our daily lives, questions are starting to be raised about the data they learned from. These questions range from potential bias or misinformation LLMs could retain from their training data to questions of copyright and fair use of human-generated text. However, while these questions emerge, developers of the recent state-of-the-art LLMs become increasingly reluctant to disclose details on their training corpus. We here introduce the task of document-level membership inference for real-world LLMs, i.e. inferring whether the LLM has seen a given document during training or not. First, we propose a procedure
for the development and evaluation of document-level
membership inference for LLMs by leveraging commonly used data sources for training and the model release date. We then propose a practical, black-box method to predict document-level membership and instantiate it on OpenLLaMA-7B with both books and academic papers. We show our methodology to perform very well, reaching an AUC of 0.856 for books and 0.678 for papers (Fig.~\ref{fig:AUC_primary}). We then show our approach to outperform the sentence-level membership inference attacks used in the privacy literature for the document-level membership task. We further evaluate whether smaller models might be less sensitive to document-level inference and show OpenLLaMA-3B to be approximately as sensitive as OpenLLaMA-7B to our approach. Finally, we consider two mitigation strategies and find the AUC to slowly decrease when only partial documents are considered but to remain fairly high when the model precision is reduced. Taken together, our results show that accurate document-level membership can be inferred for LLMs, increasing the transparency of technology poised to change our lives.\footnote{While the results we report are technically correct, recent research indicates that the high MIA performance observed might not be due to LLM memorization but rather results from a distribution shift in the collected member and non-member data. For more details, we refer to our recent results~\cite{meeus2024inherent}.}

\end{abstract}

\section{Introduction}

Over the last year, Large Language Models (LLMs) have become ubiquitous. By understanding and producing coherent natural language, models such as GPT-2/3/4~\cite{radford2019language,brown2020language,gpt4techreport}, BLOOM~\cite{scao2022bloom} and LLaMA 1/2~\cite{touvron2023llama,touvron2023llama2}, promise to revolutionise society. ChatGPT, a fine-tuned version of GPT-3, was the fastest consumer-focused application in history to reach 100 million users~\cite{chatgptfastest}. Since this breakthrough, investment in Artificial Intelligence (AI) is estimated to reach \$200 billion globally by 2025~\cite{aiinvestmentforecast}. 

While these models undoubtedly represent a major technical achievement, their capabilities stem from having been trained on enormous amounts of human-generated text. For instance, Meta's first generation model, LLaMA, had reportedly been trained on as many as 1.4 trillion tokens~\cite{touvron2023llama}. The capabilities of these models furthermore seem, at least at the moment, to keep improving with the size of the model (up to 100+ billion parameters)~\cite{bender2021dangers,kaplan2020scaling}. This--in turn--fuels the need for increasingly more data, even leaving some wondering whether they might soon have consumed all the available data on the internet~\cite{villalobos2022will}. 

\begin{figure}[t]
\centering
\includegraphics[width=0.49\linewidth]{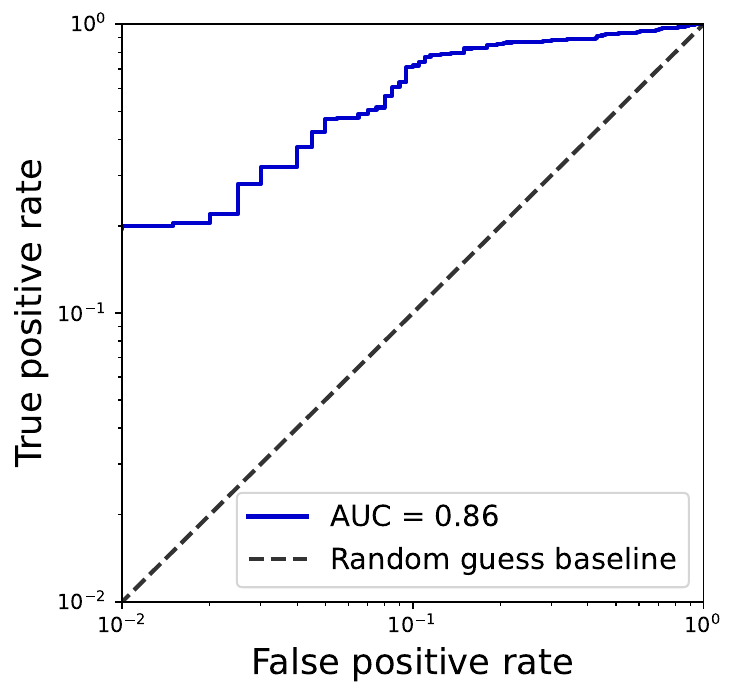} 
\includegraphics[width=0.49\linewidth]{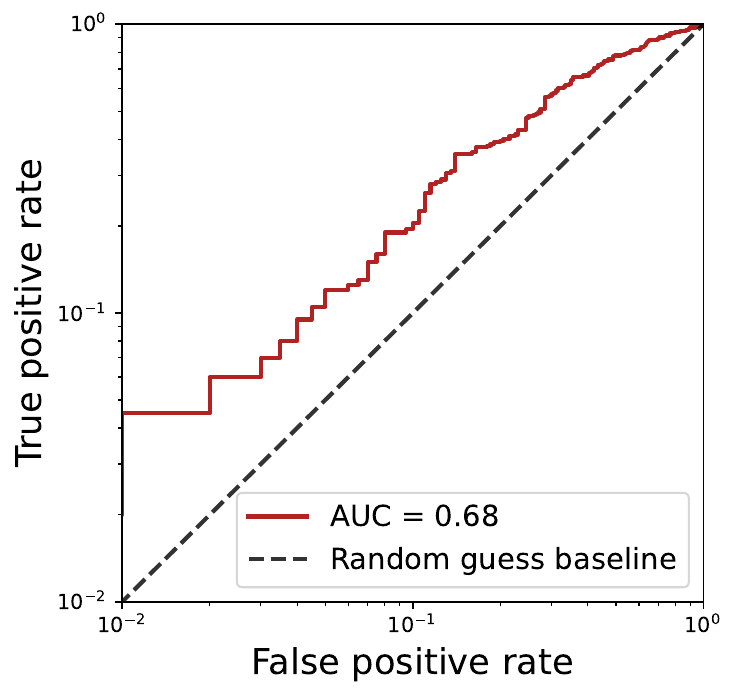} 
    \caption{ROC curve for the best performing membership classifier (see Tables~\ref{tab:books_primary} and~\ref{tab:arxiv_primary} for details). Results for books from Project Gutenberg (left) and ArXiv papers (right).} 
\label{fig:AUC_primary}
\end{figure}

As LLMs become increasingly embedded into our daily life, simplifying tasks, generating value, but also making decisions, questions are being raised regarding the data from which their capabilities emerge. This includes questions about what they will learn, e.g. whether modern models will learn the societal biases present in the data or amplify them~\cite{sheng2019woman,nadeem2020stereoset,abid2021persistent} or whether they might learn and propagate disinformation~\cite{zhang2023siren,bloombergminformation,barnard2023self,nytimeschatgptdisinformation}. Additionally, it also raises questions on ownership of the knowledge extracted by LLMs and whether copyright or \emph{fair use} apply~\cite{samuelson2023generative}. Indeed, content creators have raised concerns and filed lawsuits claiming copyright infringement by models trained on the books from the Books3 dataset, which would contain pirated copies of their content~\cite{silvermanmeta,authorsguild}, but also on songs~\cite{anthropic} or news articles~\cite{nytimes}.

Unfortunately, as those questions are being asked, model developers are becoming increasingly reluctant to disclose and discuss their training data sources. One of the currently most widely used models, GPT-4, releases no information on its training dataset~\cite{gpt4techreport}, while Meta initially released details on the training dataset of LLaMA~\cite{touvron2023llama}, but now resists to do so in their release of LLaMA-2~\cite{touvron2023llama2}. 

\textbf{Contribution.} We here introduce the concept of document-level membership inference and propose what is -to the best of our knowledge- the first setup and methodology to infer whether an LLM has seen a document during training or not.  Our method can help organizations, regulators, content creators and model users to evaluate whether a piece of content has been included in the training dataset of an LLM. 

First, we formalize the task of document-level membership inference and propose a procedure and dataset for the development and the evaluation of document-level membership inference methodologies. The development of a meta-classifier,  requires a dataset with \emph{members}, documents that were likely seen by the model during training, and \emph{non-members}, documents that are unlikely to have been seen by the model. Finding members is typically easy as most LLMs today have seen sources such as Common Crawl~\cite{commoncrawl} or Project Gutenberg~\cite{projectgutenberg}. Given the large amount of data seen by LLMs today, finding \emph{non-members} is more challenging. We thus propose to use documents similar to the one available in public sources but now made available after the LLM was released. We then construct a dataset to train and validate our inference methodology for books (Project Gutenberg~\cite{projectgutenberg}) and academic papers (ArXiv) and for OpenLLaMA~\cite{openlm2023openllama}, whose training dataset is known, RedPajama-Data~\cite{together2023redpajama}.

Next, we introduce our methodology to infer, based on black-box access to the language model, whether a given document has been seen by a model or not. The methodology consist of the following steps: querying the model for token-level predictions, normalizing the predictions for how common the token is, aggregating the predictions to the document level, and lastly building a meta-classifier.

Our approach achieves an AUC of 0.856 and 0.678 for books and  ArXiv papers respectively, on average. We also find that for this setup we retrieve a 64.4\% and 19.5\% true positive rate at a 10\% false positive rate, for books and papers respectively, implying that our classifier can accurately identify members. Given the immense corpus used to train a state-of-the-art LLM such as OpenLLaMA~\cite{openlm2023openllama}, we find it remarkable that our auditor can distinguish the contribution of an individual document. Our results also show that normalizing by both how common a token is and by the maximum probability returned by the model, is essential and that the signal for document-level membership lies in the detailed distribution of the token-level information within a document.

We then implement a sequence-level membership inference approach, such as previously considered in the privacy literature~\cite{yeom2018privacy,mattern2023membership,carlini2021extracting}. We show that state-of-the-art sequence-level baselines, scaled to document-level in three distinct ways, to  perform poorly on our setup, reaching a maximum AUC of 0.56 and 0.57 for books and papers respectively.

Further, we investigate whether smaller models would impact the ability of an auditor to infer document membership. We find that even the smallest OpenLLaMA model, with 3 billion parameters, memorizes enough information about specific documents for the inference to be successful. In fact, our results show that the AUC remains at 0.86 and 0.68 for books and ArXiv papers respectively when switching from the 7B parameters to the 3B parameters models, indicating that even "smaller" models are at risk. 

Finally, we consider potential membership inference mitigation strategies. We find that the AUC slowly decreases for partial documents, reaching an AUC of 0.62 for books when only 100 tokens are considered. For quantized versions of the language model, we still recover an AUC of 0.84 for books when the model is queried with \emph{int4} precision.


\section{Background}

\subsection{Language modeling}

Recently, large language models (LLMs) have dominated the research in natural language processing~\cite{radford2019language,brown2020language,vaswani2017attention}. Being trained in a self-supervised way on a large set of human generated text, LLMs are increasingly able to understand and generate natural language.

LLMs require a tokenizer $T$ that discretizes a sequence of text to a sequence of $n$ tokens $\{t_1,\ldots,t_n\}$. A token can be any sequence of characters that appear in language (e.g. a single word, part of a word or a combination of words). There is a finite number of tokens, and the collection of the tokens is referred to as the vocabulary $\mathcal{V}$ of size $V =|\mathcal{V}|$.

Given a sequence of $n$ tokens, "foundation" language models are trained to optimize the probability for the model to generate a sequence of tokens $\{t_1,\ldots,t_n\}$, i.e. $p(t_{1},\ldots,t_{n})$. 
More specifically, auto-regressive language models like GPT-2 and GPT-3~\cite{radford2019language,brown2020language} are trained for "next-token prediction", i.e. to predict the likelihood of the next-token given the preceding tokens. These models have shown to improve understanding and particularly generation of natural language. They compute the probability of a sequence of tokens from a series of conditional probabilities as $p(t_{1},\ldots,t_{n}) = \prod_{i=1}^{n} p(t_i | t_1,\ldots,t_{i-1})$.

The preceding sequence of tokens used to predict the probability of the next token is commonly referred to as the $\emph{context}$, consisting of length $C = |\text{context}|$. Typically, language models have a maximum context length that can be taken into account, which we denote as $\maxcontextlength$. 

LLMs use neural networks to estimate this probability distribution, with a significant amount of parameters $\theta$ to be fit during training (up to 100+ billion). The predicted probability of $\languagemodel$ with parameters $\theta$ for token $t_i$ and context of length $C$ can be formalized as $\languagemodel_{\theta}(t_i | t_{i-C}, \ldots, t_{i-1})$.

The values for $\theta$ are optimized to maximize the likelihood of the sequence of tokens in a large dataset $\trainingdata$. $\trainingdata$ can for instance include a book or a Wikipedia page and in practice contains up to trillions of tokens. Here, we denote the total number of tokens in the training data as $N_\text{train}$. More formally, the model parameters $\theta$ are determined to minimize the following loss function: 
$\mathcal{L}(\theta) = -\log \prod_{i=1}^{N_\text{train}} \languagemodel_{\theta}(t_i | t_{i-C}, \ldots, t_{i-1})$.

In the past, recurrent neural networks were the standard choice for the architecture of language models, but, over recent years, they have been replaced by attention-based models~\cite{vaswani2017attention}, in particular the transformer-based models, which now dominate the scene. 

Since the release of ChatGPT, a fine-tuned version of GPT-3 optimized for chat-like behaviour, the development of ever better language models has further accelerated. Examples include Palm 1/2~\cite{chowdhery2022palm,anil2023palm}, BLOOM~\cite{scao2022bloom}, LLaMA 1/2~\cite{touvron2023llama,touvron2023llama2} and Mistral 7B~\cite{jiang2023mistral}. At the same time, newly developed models are increasingly made publicly available on the open platform hosted by Hugging Face, which also deploys a LLM leaderboard to track and compare performances~\footnote{\href{https://huggingface.co/spaces/HuggingFaceH4/open_llm_leaderboard}{Hugging Face leaderboard}}.

\subsection{Datasets used for training}

State-of-the-art LLMs consist of billions of parameters, which are trained on large-scale datasets~\cite{bender2021dangers,kaplan2020scaling} containing trillions of tokens. Web-scraped data from the internet has long been the primary source for these large-scale datasets. Data retrieved from the frequently updated Common Crawl~\cite{commoncrawl} has for instance been the majority of the training data for GPT-3~\cite{brown2020language}, LLaMA~\cite{touvron2023llama} and BLOOM~\cite{scao2022bloom}. 

Controlling the quality of the language included from these sources is however crucial. This led to the creation of curated datasets, such as WebText (which only contains text scraped from links sufficiently recognized on the social media platform Reddit, used for GPT-2)~\cite{radford2019language} and C4~\cite{raffel2020exploring} and the inclusion of more moderated content such as from Wikipedia. 

Technology companies and researchers are in competition to develop ever better performing LLMs. The search for textual data, ideally of high quality, to be used to train the models is thus crucial. Exactly for this purpose, the training datasets have often been extended by other sources of high quality text, such as academic papers from ArXiv~\cite{lewkowycz2022solving} or books from Project Gutenberg~\cite{projectgutenberg}. The latter contains thousands of English books in the public domain. 

With the same objective of releasing high quality textual data, Goa et al~\cite{song2019auditing} released the Pile, a 800GB dataset  consisting of a diverse set of English text. This includes Books3, a dataset consisting of around 200,000 books obtained from pirate sources, most of which were published in the last 20 years and are thus not free of copyright~\cite{atlantic,wired}. It is known that for instance BloombergGPT~\cite{wu2023bloomberggpt} and LLaMA~\cite{touvron2023llama} have been trained on Books3. 

\subsection{Copyright and generative AI}

Content creators, such as authors or artists, have raised concerns about the inclusion of their work in the training data of generative AI, including LLMs but also multi-modal models such as DALLE-2~\cite{ramesh2021zero} and Stable Diffusion~\cite{rombach2022high}. These concerns have led to multiple lawsuits against technology companies who have acknowledged the use of copyrighted material for training, without the consent of the creators.

For instance, Stability AI is currently defending against two lawsuits, one filed by Getty Images~\cite{gettystability} and one filed as a class action lawsuit by multiple creators~\cite{classactionstability}, both of which argue the use of copyright-protected content to train Stable Diffusion. Further, US comedian Sarah Silverman and other authors have filed lawsuits against Meta~\cite{silvermanmeta}, claiming Meta has infringed their copyrights by training LLaMA on pirated copies of their content. The US Authors Guild also published an open letter, signed by more than 15,000 authors, calling upon AI industry leaders to protect their content~\cite{authorsguild}. Since then, other content creators have filed lawsuits against LLM developers claiming copyright infringement, including the New York Times~\cite{nytimes} and Universal Music~\cite{anthropic}.

Notably, since these lawsuits and public concerns have emerged, the original data source for the Books3 dataset has been removed~\footnote{\url{https://huggingface.co/datasets/the_pile_books3}} and technology companies tend to not disclose details on the dataset used to train the latest language models any longer~\cite{touvron2023llama2,gpt4techreport,jiang2023mistral}. 

Recognizing these ongoing court cases, Samuelson~\cite{samuelson2023generative} articulates the challenges between copyright laws and generative AI. In particular, they mention that the court will need to decide whether the inclusion of in-copyright works in the training data of AI models falls under \emph{fair use}, in which case it would not be copyright-protected.  While it is still uncertain how copyright applies to generative AI, it is clear that content creators are concerned~\cite{authorsguild} while the technology will continue to evolve rapidly without necessarily taking these concerns into account~\cite{metacounterargument,openaicounterargument}. 

\subsection{Membership inference attacks}

Homer et al.~\cite{homer2008resolving} introduced Membership Inference Attacks (MIAs) to study whether the presence of a certain target individual can be inferred from a genomic DNA mixture. They used statistical tests to distinguish between mixtures including the target and mixtures including individuals randomly drawn from a reference population. 
Since then, MIAs have been widely used to evaluate the privacy risk of aggregate data releases, such as location data~\cite{pyrgelis2017knock} or survey data~\cite{bauer2020towards}.

Since then, in the privacy literature, MIAs have also been developed against Machine Learning (ML) models~\cite{shokri2017membership,truex2019demystifying,feldman2020does,carlini2022membership,hu2022membership}. Given a certain record $D$, a ML model trained on dataset $\trainingdata$, the attacker aims to infer whether the particular record was part of $\trainingdata$ or not. In many cases, the MIA setup makes additional assumptions on the access to the training data distribution and the ML model available to the attacker. 

Shokri et al.~\cite{shokri2017membership} is seen as a foundational contribution in the field of MIAs against ML models. They assume the attacker to have black-box access to the model, i.e. they can query the model for predictions for a given input, and consider various assumptions on access to the training data. In their approach, the attacker uses the shadow modeling technique enabling them to learn a decision boundary for membership through a meta-classifier. Our proposed method also leverages a similar meta-classifier, inspired by the shadow modeling technique, while adapting it to our auditing setup.

MIAs against ML models have since been extensively studied in subsequent works and have become a common tool to assess what a ML model memorizes and whether its release poses any privacy risk~\cite{hu2022membership,carlini2022membership,song2019auditing,feldman2020does,meeus2023achilles,cretu2023re}. Importantly for this paper, MIAs have also been developed against (large) language models, which we discuss in the related work (Sec.~\ref{sec:related_work}).  


\section{Auditing setup}
We consider an auditor $\attacker$ that has access to a document $D$, e.g. a book or a research article, and wants to check if $D$ was used to train a language model $\languagemodel$. Thus, if we assume $\languagemodel$ is trained using dataset $\trainingdata$, then the auditor wants to infer whether $D \in \trainingdata$ or not. 

We define document $D$ consisting of tokens $t_i$ such that $i \in \{1,...,N\}$, or

\begin{equation}
    D = \{SOS, \text{t}_1,.., t_i, ..,\text{t}_N, EOS\}
\end{equation}
where $SOS$ is start-of-sequence token, and $EOS$ is end-of-sequence token. Any such document $D$ is highly likely to consist of significantly more tokens than the maximum context length of existing language models. Thus, we further assume that $D$ consists of $N$ tokens, and $N > C_{max}$.

We further assume that the auditor $\mathcal{A}$ has:

\begin{enumerate}
    \item Query-only access to the language model $\languagemodel$ and full access to its tokenizer $T$. This means that the auditor can query the model with a sequence of tokens $S$, and receive the probability output for all tokens $v$ in vocabulary $\mathcal{V}$. This is a realistic assumption as trained models (along with their tokenizers) like LLaMA-2~\cite{touvron2023llama2} and Mistral 7B~\cite{jiang2023mistral} are fully and freely released on platforms such as the one hosted by Hugging Face. We assume that the auditor is able to query $\languagemodel$ an arbitrary number of times.  
    \item Access to two sets of documents, that stem from the same distribution as the document $D$ e.g. if $D$ is a book then  $D_{\text{M}}$ and $D_{\text{NM}}$ also contain books. $D_{\text{M}}$ and $D_{\text{NM}}$ are defined as follows:
        \begin{enumerate}
            \item $D_{\text{M}}$: A subset of documents used in training of $\languagemodel$, also referred to as members, $\forall D \in D_{\text{M}}, D \in \trainingdata$.
            \item $D_{\text{NM}}$: A subset of documents not used in training of $\languagemodel$, also referred to as non-members, i.e. $\forall D \in D_{\text{NM}}, D \notin \trainingdata$.
        \end{enumerate}
\end{enumerate}

Note that access to $D_{\text{M}}$ and $D_{\text{NM}}$ is realistic in practice. First, most models use very similar datasets from sources such as Common Crawl~\cite{commoncrawl} and Project Gutenberg~\cite{projectgutenberg}, that are publicly available and easily accessible to include in $D_{\text{M}}$. Second, these sources are regularly updated and timestamped, which allows the auditor to construct $D_{\text{NM}}$ by collecting the data that has been added after the training data has been collected (for which reasonable assumptions can be made as the model release is typically known). We leverage exactly this intuition to build the dataset for our experiments (Sec.~\ref{sec:dataset_membership}). While not strictly required for the setup, we here consider a model for which each member document has been seen exactly once, and in its entirety, during training (Sec.~\ref{sec:exp_setup_model}). 

Further, we assume that $\languagemodel$ is a foundation model trained for next-token prediction on $\trainingdata$ which has not been subject to any measure that could shift the predicted probability distributions such as instruction tuning~\cite{zhang2023instruction}, reinforcement learning from human feedback~\cite{bai2022training} or watermarking~\cite{kirchenbauer2023watermark}. 

Under these assumptions, the auditor aims to infer whether document $D$ has been used to train the language model $\languagemodel$. 

\label{sec:auditor_model}

\section{Methodology}
The auditor $\attacker$ wants to build a meta-classifier $\metaclassifier$ able to detect whether document $D$ was used to train the language model $\languagemodel$. To achieve this, we extract document-level features for document $D$ that could carry meaningful information on the membership of documents for the training of $\languagemodel$. 

\textbf{Intuition.} From the privacy literature on MIAs against ML models~\cite{shokri2017membership,feldman2020does,yeom2018privacy}, we learn that models tend to make more confident predictions on data samples that were seen during training than on data samples that were not. It is exactly this information that an auditor can leverage to infer membership. For LLMs, this confidence is reflected in the predicted probability for the true next token in the dataset given the preceding tokens. We use this intuition to construct the document-level features, which are then used as input for our meta-classifier $\metaclassifier$. Specifically, our method considers the following four steps: 

1. We first query $\languagemodel$ to retrieve the predicted probability $\languagemodel(t_i | t_{i-C}, \ldots, t_{i-1})$ of a token given the context of length $C$, and we do this for all tokens in document $D$, i.e. $\forall t_i \in D$. We try different values of $C$ in our approach.

2. We then normalize the predicted probability of a token using a normalization algorithm $\normalization$, for which we consider different options (Sec.~\ref{sec:normalization}).

3. We then consider multiple strategies $\featureaggregation$ to aggregate all the token level information for a document $D$ to construct document level features. 

4. Finally, the meta-classifier $\metaclassifier$ takes as input the document-level features of $D$ to make a binary prediction on membership of $D$. 

We further formalize these four steps in the sections below. 

\subsection{Querying the model}
\label{sec:query_model}

The auditor queries $\languagemodel$ to retrieve a value per token $t_i$ in document $D$. As the maximum context length of the model $\maxcontextlength$ is likely to be smaller than the length of the document $N$, the auditor runs the model through the document with a certain fixed context length $C$ and stride $s$. 

The document is split in $N_s$ sequences $S_j, j = 1,...,N_s$ with $S_j = (t_{j,1}, \ldots, t_{j,C})$ consisting of $C$ consecutive tokens. The last sequence has $C'$ tokens, which can be $\leq C$ depending on the total number of tokens $N$ in $D$ and stride $s$. We apply the same operations for this last sequence using $C'$ instead of $C$. 

Each sequence is passed to the model as input and results in a sequence of predicted probabilities for the corresponding true tokens:

\begin{align}
\begin{split}
\languagemodel_{\theta}(S_j) = & \left\{\languagemodel_{\theta}(t_{j,2} | t_{j,1}),\ldots \languagemodel_{\theta}(t_{j,C} | t_{j,1} \ldots t_{j,C-1}) \right\}
\end{split}
\end{align}

Note that $|\languagemodel_{\theta}(S_j)| = |S_j| - 1 = C -1$, as the model does not return a prediction for the first token $t_{j,1}$ in the absence of a meaningful context to do so. Thus, when using context length $C$, we move through the document with stride $s = C-1$. We then get sequences $S_j$ for $j = 1 \ldots{N_s}$ where $N_s = \lceil\frac{N}{C-1}\rceil$, so the resulting set of all $\languagemodel_{\theta}(S_j)$ contains predictions for all tokens $t_i$ in $D$, except for the very first token $t_1$. We further refer to this probability of token $t_i$ in document $D$ for language model $\languagemodel_{\theta}$ simply as $\languagemodel(t_i)$.

The value for $C$ is a fundamental hyperparameter of the setup and can be chosen to be any integer value smaller than $\maxcontextlength$. Note that the predictions of a language model typically become more accurate for a longer context length $C$. However, as the goal is to retrieve information that should be meaningful for membership, it is not clear if this information is more likely to be at predictions with smaller or larger context.

\subsection{Normalization (\texorpdfstring{$\normalization$}{Lg})}
\label{sec:normalization}

In the section above, we query $\languagemodel$ to extract $\languagemodel(t_i)$, i.e. the probability with which the $\languagemodel$ would predict the next token in question given a certain context. 
However, in natural language, some tokens are more rare than others. Intuitively, the probability with which a model predicts a certain token would depend on how rare such a token is. For instance, for a frequently occurring token such as \emph{the} or \emph{and} the model might predict a high probability, while for a more infrequent token the model naturally predicts a lower probability. We hypothesize that this occurs regardless of whether the model has seen the sequence of interest at training time or not. 

Thus, to optimally retrieve the information meaningful for membership we \emph{normalize} the model output with a value that takes into account the rarity of a token.
Prior work on sequence-level membership inference has approached this in various ways. For instance, Carlini et al.~\cite{carlini2021extracting} consider the zlib entropy of a sequence or use the predicted probability of a reference model. While zlib considers the entropy of a sequence and does not provide token-specific information, the use of a reference model requires using another model which also uses the exact same tokenizer $T$ as $\languagemodel$, which is often not available. Instead, we propose different token-specific $\normalization$ approaches based on the data and the model the auditor $\attacker$ already has access to.

\subsubsection{Computing a normalization dictionary}
\label{sec:norm_dict}

We consider the token $t_i=v$ in document $D$ with $v$ the token value part of the vocabulary $\vocabulary$. We then normalize the predicted probability of $t_i$, $\languagemodel(t_i)$, using reference value $R(v)$. $R(v)$ is calculated for each token $v$ in vocabulary $\vocabulary$ to capture the rarity of $v$. We propose two ways of computing $R(v)$.

\textbf{Token frequency (TF).} Recall that the auditor is assumed to have access to both $D_{\text{M}}$ and $D_{\text{NM}}$. With the concatenated dataset as reference $D_{\text{ref}} = D_{\text{M}} \cup D_{\text{NM}}$, the token frequency ${R^{TF}(v)}$ for token $v$ in vocabulary $\vocabulary$ is then computed as follows:

\begin{align}
{R^{TF}(v)} = \frac{\text{count}(v, D_\text{ref})} {\sum_{j=1}^{V} \text{count}(v_j,D_\text{ref})}
\end{align}

where $\text{count}(v, D_\text{ref})$ corresponds to the number of times token $v$ appears in the set of documents $D_\text{ref}$. For tokens that do not appear even once in our dataset $D_{\text{ref}}$ we use the smallest frequency in the normalization dictionary divided by 2 as a reference value.

\textbf{General probability (GP).} While the token frequency quantifies how rare a certain token is in the dataset, its computation does not take into account the specific context with which the token appears. Also, when a token $v$ does not appear even once in $D_\text{ref}$, no valid value for the frequency ${R^{TF}(v)}$ can be computed. 

In order to address both concerns, we propose to compute the \emph{general probability} $R^{GP}(v)$ of the token $v$ in vocabulary $\vocabulary$. Here, we run through $D_\text{ref}$ as described in Sec.~\ref{sec:query_model} and compute the average of all the predicted probabilities for $v$. Note that we consider all predictions, also when the true token $t_k \in D_\text{ref}$ is not equal to $v$, as the model predicts the probability distribution over the entire vocabulary every time. $R^{GP}(v)$ is computed as follows: 

\begin{align}
R^{GP}(v) = \frac{1}{|D_\text{ref}|} \sum_{k=1}^{|D_\text{ref}|} \languagemodel_{\theta}(v \mid t_{k-C}, \ldots, t_{k-1})
\end{align}

\subsubsection{Normalization strategies}

We here propose different normalization strategies $\normalization$ for $\languagemodel(t_i=v)$ using $R(v)$.

\textbf{No normalization (\textit{NoNorm}).} In this case we use the predicted probabilities as they are and do not apply any normalization, or: 

\begin{align}
    \textit{NoNorm}(t_i=v) = & \languagemodel(t_i)
\end{align}

\textbf{Ratio normalization using the token frequency (\textit{RatioNormTF}).} We here compute the ratio of the predicted probability of the token of interest and the corresponding token frequency $R^{TF}(v)$ in $D_\text{ref}$ as discussed in Sec. \ref{sec:norm_dict}.

\begin{align}
\textit{RatioNormTF}(t_i=v) =  \frac{\languagemodel(t_i)}{R^{TF}(v)}
\end{align}

\textbf{Ratio normalization using the general probability (\textit{RatioNormGP}).} We here compute the ratio of the predicted probability of the token of interest and the corresponding general probability $R^{GP}(v)$ in $D_\text{ref}$ as discussed in Sec. \ref{sec:norm_dict}. 

\begin{align}
\textit{RatioNormGP}(t_i=v) =  \frac{\languagemodel(t_i)}{R^{GP}(v)}
\end{align}

\begin{figure}[t]
\centering
\includegraphics[width=0.7\linewidth]{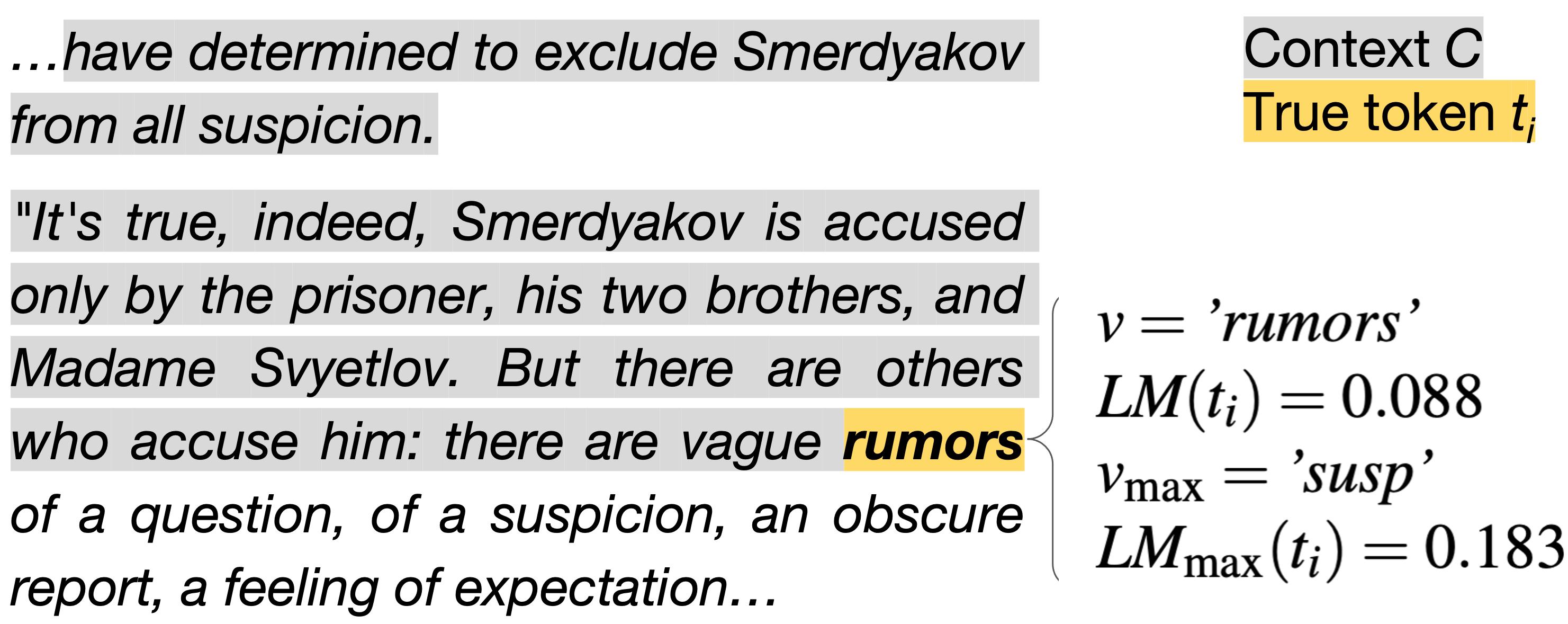} 
    \caption{Querying OpenLLaMA-7B on an example from the book \emph{The Brothers Karamazov by Dostoyevsky} (member, Sec.~\ref{sec:dataset_membership}).}
\label{fig:example_sentence}
\end{figure}

\textbf{Maximum probability normalization (\textit{MaxNorm}).} For a given token $t_i$, the model predicts the probability distribution over all tokens in the vocabulary $\mathcal{V}$. For now we have only considered the probability that the model returns for the true token $t_i$, regardless of how this probability compares to the predicted probability for the other token values with the same context. Intuitively, we could expect that the difference between the maximum predicted probability over all tokens $t \in \vocabulary$, with corresponding token $v_{\text{max}}$, and the predicted probability for the true token $t_i$ carries information about how likely the model is to predict the token of interest. Fig.~\ref{fig:example_sentence} illustrates how the predicted probability for the true token and the maximum probability differ on a real example. We hypothesize that exactly the difference between both values could be meaningful to infer membership. 

Formally, we denote the maximum probability being predicted for $t_i$ for context $C_i$ as 

\begin{equation}
    \languagemodel_{\text{max}}(t_i|C_i) = \max_{t \in \vocabulary}\languagemodel(t|C_i)
\end{equation}

regardless of whether the probability corresponds to the true token $t_i$ or not. For rest of the paper, $\languagemodel_{\text{max}}(t_i)$ implies $\languagemodel_{\text{max}}(t_i|C_i)$. We then combine this with the ratio normalization strategies to get $\textit{MaxNormTF}$ and $\textit{MaxNormGP}$.

\begin{equation}
    \textit{MaxNormTF}(t_i=v) = \frac{ 1 - (\languagemodel_{\text{max}}(t_i) - \languagemodel(t_i))}{R^{TF}(v)}
\end{equation}

\begin{equation}
    \textit{MaxNormGP}(t_i=v) = \frac{ 1 - (\languagemodel_{\text{max}}(t_i) - \languagemodel(t_i))}{R^{GP}(v)}
\end{equation}

Note that we ensure that the numerator never equals zero, even when the model predicts the highest probability for the true token $t_i$, or when $\languagemodel_{\text{max}}(t_i) =\languagemodel(t_i)$.

\subsection{Feature extraction (\texorpdfstring{$\featureaggregation$}{Lg})}

For all the tokens $t_i$ in document $D$, we have so far normalized the predicted probabilities using normalization strategy $\normalization$. We now use these normalized values to set $F(t_i)$ as the negative log of the output after normalizing token-level probabilities, or $F(t_i)=-\log(\normalization(\languagemodel(t_i)))$.

Note that when we do not apply any normalization, or \textit{NoNorm}, the value for $F(t_i)$ corresponds to the cross-entropy loss for the predicted probability, as used to optimize the model parameters during training. 

While we now have computed a final value on the token level, our ultimate goal remains to predict binary membership for a document. Hence, from all token-level information $F(t_i), \forall t_i \in D$, we need to extract document-level features that could capture meaningful information for membership. We consider two feature extractors \textit{AggFE} and \textit{HistFE}.

\textbf{Aggregate feature extractor (\textit{AggFE}).} For each document $D$, \textit{AggFE} computes aggregate statistics from the token-level information. Specifically, it uses $F(t_i), \forall t_i \in D$ to compute the minimum, maximum, mean and standard deviation and the values at $x$-percentiles for $x \in X_\text{perc}$.  

\textbf{Histogram feature extractor (\textit{HistFE}).} \textit{HistFE} uses $N_b$ equal-sized histogram bins and for each document it computes the fraction of total values $F(t_i)$ in each bin. These $N_b$ features are then used as document-level features. The $N_b$ equal-sized bins are determined using all token-level values in the training dataset, across all documents. 

\subsection{Meta-classifier}

As a final step, the document-level features extracted using $\featureaggregation$ are used as input to the meta-classifier $\metaclassifier$, which returns a prediction for binary membership of input document $D$. As $\metaclassifier$ we only consider a random forest classifier, allowing to fit non-linear dependencies across input features. We train $\metaclassifier$ on a training subset of $D_{\text{M}}$ and $D_{\text{NM}}$ and evaluate its performance for binary classification on a disjoint subset. 

\label{sec:methodology}

\section{Experimental setup}
\subsection{Model}
\label{sec:exp_setup_model}

As language model $\languagemodel$ we use OpenLLaMA \cite{openlm2023openllama}, a fully open-source reproduction of LLaMA~\cite{touvron2023llama}, an auto-regressive LLM developed by Meta. While LLaMA is made publicly available for research and details on their training data have been provided ~\cite{touvron2023llama}, the exact training dataset has not been released. Instead, OpenLLaMA is trained on RedPajama-Data~\cite{together2023redpajama}, which is a best-guess, open replication of the original dataset used to train LLaMA. We here opt for OpenLLaMA in order to be in full control of the training dataset. 
The models are publicly available hosted by Hugging Face in three sizes: 3, 7 and 13 billion parameters. The tokenizer $T$ has a vocabulary of size $V = 32,000$ and the maximum input length equals $\maxcontextlength = 2048$. 

\subsection{Dataset for membership}
\label{sec:dataset_membership}

We use two distinct types of documents, books and academic papers in LaTeX, to evaluate the effectiveness of our methods in inferring the document-level membership. For each type, we collect a fixed number of member ($D_{\text{M}}$) and non-member ($D_{\text{NM}}$) documents and train a separate meta-classifier. 

As indicated in Sec.~\ref{sec:auditor_model}, we use realistic assumptions for the auditor to collect both $D_{\text{M}}$ and $D_{\text{NM}}$ for each document type. Specifically, for the members, we use data sources that are typically used for language modeling and are readily available. For non-members, we leverage the fact that these sources are regularly updated to retrieve the documents that have been added after the time the training data has been collected. While we believe these assumptions are reasonable in practice, we additionally ensure there is no overlap by using our exact knowledge of RedPajama-Data~\cite{together2023redpajama}. Below we describe in detail how we approach this for each document type. 

\subsubsection{Books (Project Gutenberg)}
\label{exp_setup_books}

Project Gutenberg~\cite{projectgutenberg} is a volunteering project aiming to digitize and freely distribute books. The resulting online library contains thousands of books from a wide range of languages and publication dates. The vast majority of the releases are in the public domain in the United States, meaning that copyright laws do not apply. 

\textbf{Members $D_{\text{M}}$}. We note that the RedPajama-Data includes both books from Project Gutenberg and the Books3 dataset made available by the Pile~\cite{gao2020pile}. We exclusively consider the set of books PG-19 made available by Rae et al.~\cite{rae2019compressive}. This dataset contains 28,752 English books that were published before 1919 and made available on Project Gutenberg.

\textbf{Non-members $D_{\text{DM}}$}. Project Gutenberg is an ongoing project with new books added regularly. We create a comparable book dataset that has not been used for training by downloading books added to Project Gutenberg after the PG-19 dataset was created. Of all books included in PG-19, the latest release date on the Gutenberg project was February 10, 2019. We then use an open source library~\cite{kpullygutenberg} to download all English books that were added to Project Gutenberg after this date. In our setup, this led to a total of 9,542 books that we could use as non-members. In line with how PG-19 has been constructed, we only consider the text between the explicit start and end of the uniformly formatted text files. 

\begin{figure}[!t]
\centering
\includegraphics[width=0.49\linewidth]{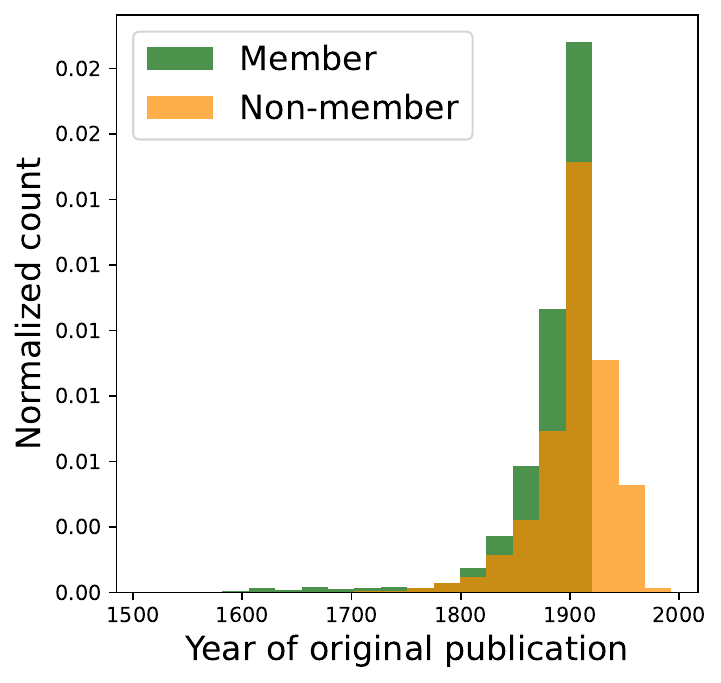} 
\includegraphics[width=0.49\linewidth]{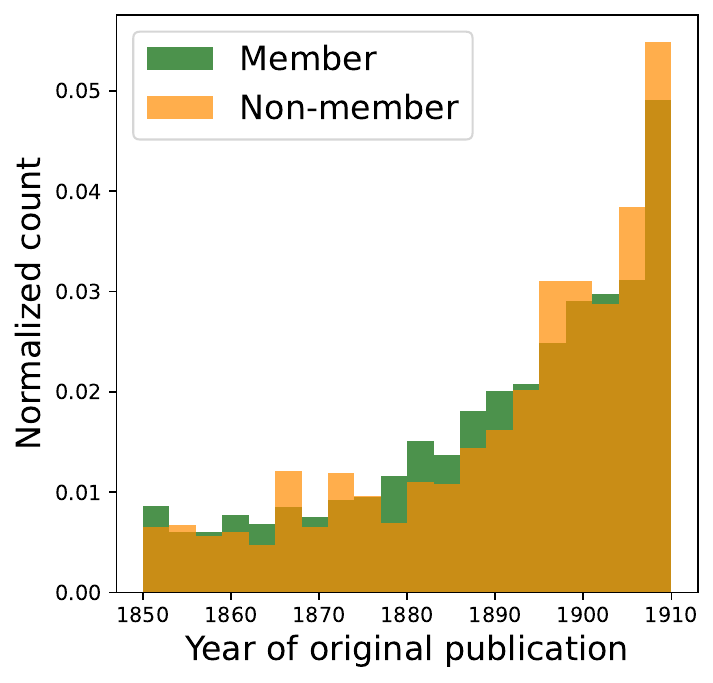} 
    \caption{Density distribution of the original year of publication for books included in Project Gutenberg, for members and non-members. Both the raw distribution (left) and the filtered distribution for years 1850-1910 (right) are displayed.} 
\label{fig:book_dates}
\end{figure}

Two books published in different eras could be easily distinguished from the writing style. In that regard, Fig.~\ref{fig:book_dates} shows that there is a meaningful shift in year of original publication between books considered as member and non-members. Thus, there is a possibility that language written in books from for instance the 1700s can lead to data drift compared to books written in the 1900s. To ensure the meta-classifier $\metaclassifier$ focuses on the memorization of the model, and does not focus on a potential drift in language, we only consider books with a year of original publication between 1850 and 1910. Fig.~\ref{fig:book_dates} shows that the distributions of members and non-members for these now filtered books are highly similar. This makes us confident that the model would focus on the membership rather than on the language drift.

\subsubsection{Academic papers (ArXiv)}

We use academic papers as posted on the online research-sharing platform ArXiv. In line with prior work training LLMs on academic papers~\cite{lewkowycz2022solving}, RedPajama-Data includes the raw LaTeX text after removing the bibliography, the comments and everything before the occurrence of the first section. 

\textbf{Members $D_{\text{M}}$.} As RedPajama-Data includes the preprocessed LaTeX text from ArXiv papers, we download documents that were part of the training dataset using their instructions~\footnote{\href{https://huggingface.co/datasets/togethercomputer/RedPajama-Data-1T}{Hugging Face RedPajama Documentation}}. This leads to approximately 1.5 million ArXiv papers to be considered as members of the training dataset.

\textbf{Non-members $D_{\text{NM}}$.} The Github repository of RedPajama-Data~\cite{together2023redpajama} also provides instructions on how to download academic papers from ArXiv directly. ArXiv maintainers host an Amazon S3 bucket containing all source LaTeX files, which gets updated on a monthly basis and can be retrieved at a moderate charge~\footnote{\url{https://info.arxiv.org/help/bulk_data_s3.html}}. The data is grouped by the month on which the first version of the document has been added to ArXiv. From RedPajama-Data, we found that the last month that had been included in the training dataset was February 2023, which allows us to download all academic papers uploaded on ArXiv from March 2023 and use them as non-members. After applying the same preprocessing recipe as for the members, our non-member dataset of ArXiv paper eventually consists of around 155 thousand documents. 

\subsubsection{Training and evaluation setup}
\label{sec:data_split}

We use the documents considered as members and non-members to train and validate our method to predict membership of document $D$ in the training data. 
To ensure our model focuses on documents of a reasonable length, and sufficiently longer than the maximum context length, $\maxcontextlength$, we remove all documents, from both the books and academic papers, that have fewer than 5,000 tokens. We then randomly sample 1,000 member and 1,000 non-member documents. 
Next, we construct $k=5$ different chunks of $400$ documents, each chunk consisting of $200$ member documents and $200$ non-member documents. This allows us to train our set-up using $k-1$ chunks, i.e. $1600$ documents, and validate the performance on the held-out chunk, i.e. on $400$ documents. By executing this $k$ times, the classification performance is robustly estimated with a mean and standard deviation across $k$ runs.
For both document types, books and academic papers, we construct this setup to train and evaluate a distinct meta-classifier $\metaclassifier$ $k$ times.

\subsubsection{Methodology parameters}
\label{sec:meth_params}

We here describe the methodology parameters (Sec.~\ref{sec:methodology}) used in our experiments. 

As context length $C$ to query the model, we consider three different values$\{128, 1024,2048\}$, the last of which is same as the maximum context length $\maxcontextlength$. As $D_\text{ref}$ to compute the normalization dictionaries in $\normalization$, we use the same sets of documents used to train the meta-classifier. This means that $k-1 = 4$ chunks, with a total $800$ members and $800$ non-members are used, leading to $|D_\text{ref}| = 1,600$ for every run. For feature extractor \textit{AggFE} we use a set of percentiles $X_\text{perc} = \{1, 5, 10, 25, 50, 75, 90, 95, 99\}$ and for \textit{HistFE} number of bins $N_b = 1,000$. For the meta-classifier $\metaclassifier$ we use a random forest classifier from scikit-learn~\cite{scikit-learn} with 500 trees of a maximum depth of 5 and a minimum of samples per leaf of 3. We ran all language model queries on a set of A100 NVIDIA GPUs with a floating point precision of 16 bits.


\section{Results}
\subsection{Evaluation across setups and datasets}
\label{sec:results_primary}

Tables~\ref{tab:books_primary} and~\ref{tab:arxiv_primary} summarize the performance of our binary membership classifier for books from Project Gutenberg and ArXiv papers respectively. We measure the performance of $\metaclassifier$ using the Area Under the Receiver Operating Characteristic Curve (AUC-ROC) and report the mean and standard deviation over $k=5$ runs (Sec.~\ref{sec:data_split}). 

\begin{table*}
    \centering
    \caption{Books from Project Gutenberg - mean and standard deviation AUC for binary membership across $k$ folds.}    \label{tab:books_primary}
    \begin{tabular}{clccc}
    \toprule
         & & \multicolumn{3}{c}{Context length $C$} \\
        $\featureaggregation$ & $\normalization$ & 128 & 1024 & 2048 \\
        \midrule
        \textit{AggFE} & \textit{NoNorm} & $0.550 \pm 0.011$ & $0.551 \pm 0.009$ & $0.554 \pm 0.012$ \\ 
        \cmidrule{2-5}
         & \textit{RatioNormTF} & $0.605 \pm 0.025$ & $0.552 \pm 0.017$ & $0.556 \pm 0.012$ \\ 
         \cmidrule{2-5}
         & \textit{RatioNormGP} & $0.621 \pm 0.031$ & $0.539 \pm 0.022$ & $0.553 \pm 0.021$ \\ 
         \cmidrule{2-5}
         & \textit{MaxNormTF} & $0.620 \pm 0.029$ & $0.546 \pm 0.021$ & $0.543 \pm 0.016$ \\ 
         \cmidrule{2-5}
         & \textit{MaxNormGP} & $0.626 \pm 0.025$ & $0.542 \pm 0.025$ & $0.553 \pm 0.021$ \\ 
         \midrule
        \textit{HistFE} & \textit{NoNorm} & $0.566 \pm 0.022$ & $0.528 \pm 0.024$ & $ 0.541 \pm 0.017$ \\ 
        \cmidrule{2-5}
         & \textit{RatioNormTF} & $0.766 \pm 0.030$ & $0.786 \pm 0.028$ & $0.799 \pm 0.020$ \\ 
         \cmidrule{2-5}
         & \textit{RatioNormGP} & $0.779 \pm 0.010$ & $0.790 \pm 0.013$ & $0.804 \pm 0.011$ \\ 
         \cmidrule{2-5}
         & \textit{MaxNormTF} & \bm{$0.856 \pm 0.021$} & $0.854 \pm 0.026$ & $0.855 \pm 0.027$ \\ 
         \cmidrule{2-5}
         & \textit{MaxNormGP} & $0.849 \pm 0.019$ & $0.853 \pm 0.0218$ & $0.852 \pm 0.026$ \\ 
         \bottomrule
    \end{tabular}
\end{table*}

The highest mean AUC achieved across setups for books and papers is $0.856$ and $0.678$ respectively, which is significantly higher than the random baseline of $0.5$. This effectively shows that our model queries, normalization strategies and feature extraction enables the meta-classifier to learn the decision boundary for membership of a document $D$ in the training data of an LLM. Around 250,000 books and 1.5 million ArXiv papers  contribute to around 4.5\% and 2.5\% of the entire training dataset~\cite{touvron2023llama}. This makes the contribution of one document to the entire dataset negligible, and highlights the difficulty of membership inference. In this context, our meta-classifier's performance demonstrates the ability of our simple $\normalization$ and $\featureaggregation$ strategies to separate signal from the noise.

Further, the results allow us to compare the combinations of different setups, i.e. combination of context length $C$, normalization strategy $\normalization$, and feature aggregation strategy $\featureaggregation$ as discussed in Sec.~\ref{sec:methodology}. First, we notice fairly little difference in performance across different values of context length $C$. This would imply that memorization is equally exposed for lower and larger values of context length. Second, we find that normalization is required to reach reasonable classification performances. Compared to the raw predicted probabilities \textit{NoNorm}, all normalization strategies lead to an improved performance. Notably for \textit{MaxNormTF}, we find that, for $C=128$ and \textit{HistFE}, the mean AUC increases by $0.29$ and $0.10$ for books and papers respectively when compared with \textit{NoNorm} performance. Additionally, we find fairly little difference between normalizing with the \emph{token frequency} $R^{TF}(v)$ or \emph{general probability} $R^{GP}(v)$ across setups. This would imply that both normalization dictionaries provide similar information on the rarity of a token value. Third, we find that the histogram feature extraction \textit{HistFE} is significantly more effective than the more simple aggregate statistics extraction \textit{AggFE}. For \textit{MaxNormTF} and $C=128$, \textit{HistFE} leads to an increase in AUC of $0.24$ and $0.05$ for books and papers respectively compared with \textit{AggFE} in an equivalent setup. This makes us confident that the information for membership truly lies in the detailed distribution of token-level values per document, rather than in high-level aggregates. Overall, $C=128$, normalization using \textit{MaxNormTF}, and feature aggregation using \textit{HistFE} results in the best performing meta-classifier $\metaclassifier$ for both books and academic papers.

\begin{table}[ht]
    \centering
    \begin{tabular}{ccc}
    \toprule
         & \multicolumn{2}{c}{TPR@FPR} \\
        Dataset & 10\% & 1\% \\
        \midrule 
         Project Gutenberg & $64.44 \pm 9.32 \%$ & $18.75 \pm 3.82 \%$\\ 
        \cmidrule{2-3}
        ArXiv & $19.50 \pm 0.94 \%$ & $5.92 \pm 2.13 \%$ \\ 
        \bottomrule
    \end{tabular}
    \caption{True positive rates at low false positive rates for the best setup in Tables~\ref{tab:books_primary} and~\ref{tab:arxiv_primary}.}
    \label{tab:tpr_fpr}
\end{table}

Fig.~\ref{fig:AUC_primary} shows the ROC curve for one trained meta-classifier $\metaclassifier$, randomly selected out of the $k$, for the best performing setup. In line with Carlini et al.~\cite{carlini2022membership}, we also provide the true positive rate at low false positive rates for this setup in Table~\ref{tab:tpr_fpr}. Especially for the books, a true positive rate of $18.75\%$ at a false positive rate of $1\%$ implies a meta-classifier $\metaclassifier$ that can confidently identify members in a given set of documents.

\begin{table*}[ht]
    \centering
    \caption{ArXiv papers - mean and standard deviation AUC for binary membership across $k$ folds.}
    \begin{tabular}{clccc}
    \toprule
         & & \multicolumn{3}{c}{Context length $C$} \\
        $\featureaggregation$ & $\normalization$ & 128 & 1024 & 2048 \\
        \midrule
        \textit{AggFE} & \textit{NoNorm} & $0.617 \pm 0.015$ & $0.605 \pm 0.031$ & $0.602 \pm 0.022$ \\ 
        \cmidrule{2-5}
         & \textit{RatioNormTF} & $0.609 \pm 0.027$ & $0.614 \pm 0.027$ & $0.613 \pm 0.017$ \\ 
         \cmidrule{2-5}
         & \textit{RatioNormGP} & $0.605 \pm 0.028$ & $0.609 \pm 0.019$ & $0.616 \pm 0.014$ \\ 
         \cmidrule{2-5}
         & \textit{MaxNormTF} & $0.630 \pm 0.021$ & $0.622 \pm 0.026$ & $0.623 \pm 0.026$ \\ 
         \cmidrule{2-5}
         & \textit{MaxNormGP} & $0.635 \pm 0.016$ & $0.632 \pm 0.022$ & $0.626 \pm 0.023$ \\ 
         \cmidrule{2-5}
        \textit{HistFE} & \textit{NoNorm} & $0.579 \pm 0.029$ & $0.580 \pm 0.026$ & $0.580 \pm 0.028$ \\ 
        \cmidrule{2-5}
         & \textit{RatioNormTF} & $0.644 \pm 0.028$ & $0.647 \pm 0.026$ & $0.654 \pm 0.033$ \\ 
         \cmidrule{2-5}
         & \textit{RatioNormGP} & $0.643 \pm 0.031$ & $0.638 \pm 0.026$ & $0.645 \pm 0.030$ \\ 
         \cmidrule{2-5}
         & \textit{MaxNormTF} & \bm{$0.678 \pm 0.024$} & $0.668 \pm 0.031$ & $0.668 \pm 0.031$ \\ 
         \cmidrule{2-5}
         & \textit{MaxNormGP} & $0.675 \pm 0.019$ & $0.665 \pm 0.021$ & $0.668 \pm 0.029$ \\ 
         \bottomrule
    \end{tabular}
    \label{tab:arxiv_primary}
\end{table*}

\subsection{Comparison to sequence-level baseline}
\label{sec:sequence_baseline}

In the privacy literature, prior work on Membership Inference Attacks (MIAs) against language models has exclusively focused on inferring membership at the sequence-level ~\cite{carlini2021extracting,carlini2022membership,mattern2023membership,yeom2018privacy}. In contrast, our setup concerns an auditor who aims to infer document-level membership. 

We here compute how state-of-the-art sequence-level MIAs perform in our setup when scaled to the document-level. We consider as MIA methodologies to compute a sequence-level membership score $\alpha$: 

1. The \emph{Loss} attack~\cite{yeom2018privacy}, which uses the language model cross-entropy loss computed for the given sequence. 

2. Both the \emph{Zlib} attack and \emph{Lower} attack~\cite{carlini2021extracting}, which divides the target language model loss by the sentence zlib compression entropy and the target language model loss computed with all lower-case characters. 

3. The \emph{Ratio} attack~\cite{carlini2021extracting}, which divides the target language model loss for a given sample by the loss computed using a reference model. As a reference model, we here use both GPT-2~\cite{radford2019language} (\emph{Ratio-GPT-2}) and the OpenLLaMA-3B~\cite{openlm2023openllama} (\emph{Ratio-3B}).  

4. The \emph{Neighborhood} attack~\cite{mattern2023membership}. For each sample, we use a RoBERTa-based masked language model~\cite{liu2019roberta} to sample $50$ neighboring samples with $1$ token replacement. We use the difference between the target language model loss computed on the sample and the mean loss of the neighboring samples. 

We now consider whether the sequence-level MIA can be scaled to the document-level. First, we split each document $D$ in sequences $S_j = (t_{j,1}, \ldots, t_{j,C})$ of the same length $C=128$. We found 1000 tokens to approximately map to 750 words, and hence using $C=128$ tokens is a good proxy for selecting a sentence. For each document $D$, we randomly sample 40 sequences $S_j$, ensuring each document has the same number of sequences and then use the same membership label for sequences as for the corresponding document. We consider three ways of scaling the sequence-level membership inference to the level of the document:

1. \emph{Sequence}. We compute the AUC on the sequence-level membership scores directly. 

2. \emph{Document (average)}. We use the average sequence-level membership score per document to compute the document-level AUC, in line with the group-level attack from~\cite{jagannatha2021membership}.

3. \emph{Document (threshold)}. We consider a threshold $T$ for the membership score, below which a sequence is classified as member and non-member otherwise. We determine $T$ so that a maximum classification accuracy is reached on the sequence-level on the training dataset. We then take the ratio of sequences within a document, i.e. $\forall (S_j) \in D$, that are predicted as member, to compute the document-level AUC. 

Similarly to Sec.~\ref{sec:results_primary}, we compute the AUC for membership on the sequence-level for $k$ chunks. Table~\ref{tab:baselines} summarizes the results, with the mean and standard deviation AUC for all MIA methodologies and strategies to scale to the document-level. 

\begin{table*}[ht]
    \centering
    \caption{Baselines - mean and standard deviation AUC for binary membership across $k$ folds.}
    \begin{tabular}{lccc|ccc}
    \toprule
         & \multicolumn{3}{c}{Project Gutenberg} & \multicolumn{3}{c}{ArXiv} \\
         & Sequence & Document & Document & Sequence & Document & Document \\
        Attack & & (average) & (threshold) & & (average) & (threshold) \\
        \midrule
        \textit{Loss} & $0.485 \pm 0.012$ & $0.478 \pm 0.025$ & $0.533 \pm 0.016$ 
                      & $0.530 \pm 0.012$ & $0.556 \pm 0.027$ & $0.563 \pm 0.028$  \\ 
         \cmidrule{1-7}
        \textit{Zlib} & $0.453 \pm 0.018$ & $0.436 \pm 0.025$ & $0.541 \pm 0.012$ 
                      & $0.511 \pm 0.007$ & $0.522 \pm 0.015$ & $0.527 \pm 0.015$  \\ 
         \cmidrule{1-7} 
        \textit{Lower} & $0.505 \pm 0.010$ & $0.537 \pm 0.024$ & $0.520 \pm 0.024$ 
                      & $0.531 \pm 0.009$ & \bm{$0.572 \pm 0.020$} & $0.568 \pm 0.022$  \\ 
         \cmidrule{1-7}
        \textit{Ratio-GPT-2} & $0.454 \pm 0.011$ & $0.407 \pm 0.014$ & $0.500 \pm 0.009$ 
                      & $0.505 \pm 0.014$ & $0.510 \pm 0.025$ & $0.530 \pm 0.025$  \\ 
         \cmidrule{1-7}
        \textit{Ratio-3B} & $0.514 \pm 0.006$ & \bm{$0.559 \pm 0.022$} & $0.553 \pm 0.024$ 
                      & $0.485 \pm 0.006$ & $0.449 \pm 0.022$ & $0.489 \pm 0.010$  \\ 
         \cmidrule{1-7}
        \textit{Neighborhood} & $0.508 \pm 0.005$ & $0.532 \pm 0.025$ & $0.530 \pm 0.020$ 
                      & $0.506 \pm 0.010$ & $0.513 \pm 0.032$ & $0.519 \pm 0.025$  \\ 
        \cmidrule{1-7}
        \textit{Min-K\% Prob}*& $0.491 \pm 0.008$ & $0.486 \pm 0.017$ & $0.523 \pm 0.020$ 
                      & $0.531 \pm 0.011$ & $0.553 \pm 0.024$ & $0.557 \pm 0.019$  \\ 
         \bottomrule
    \end{tabular}
      \begin{tablenotes}
      \item[*] *Concurrent work.
      \end{tablenotes}
    \label{tab:baselines}
\end{table*}

Overall, we find most baselines to barely perform better than a random guess, with the maximum AUC of $0.56$ and $0.57$ reached using \textit{Ratio-3B} and \textit{Lower} for books and papers respectively. 
This suggests that sequence-level MIAs from the privacy literature might not work out-of-the-box for LLMs trained on a significantly larger corpus than previously considered~\cite{carlini2022membership,carlini2021extracting,mattern2023membership}.
Further, this implies that using sequence-level information to predict document-level membership is possibly sub-optimal. Indeed, the signal for document membership might lie within the distribution of the token-level predictions rather than in the sequence-level predictions.

\textbf{Comparison with concurrent work.} Concurrently with this work, \emph{Min-K\% Prob} has been proposed as a new technique for sequence-level membership inference, which can also be scaled to detect document-level contributions to language model pretraining data~\cite{shi2023detecting}. Table~\ref{tab:baselines} also summarizes the results for $K=20$. For both datasets, we find \emph{Min-K\% Prob} to perform barely above a random guess baseline.

\subsection{Evaluation across model sizes}
\label{sec:model-size-evaluation}

Note that all results in the section above were achieved with the OpenLLaMA model with 7 billion parameters (7B)~\cite{openlm2023openllama}. We here evaluate whether our membership inference methodology would work equally well for models with less parameters. We thus use $C=2048$ and \textit{HistFE} feature-aggregation strategy, against the model with 3B parameters. We consider the two best performing normalization strategies \textit{MaxNormTF} and \textit{MaxNormGP}, along with \textit{NoNorm} for comparison.

\begin{figure}
\centering
\includegraphics[width=0.49\linewidth]{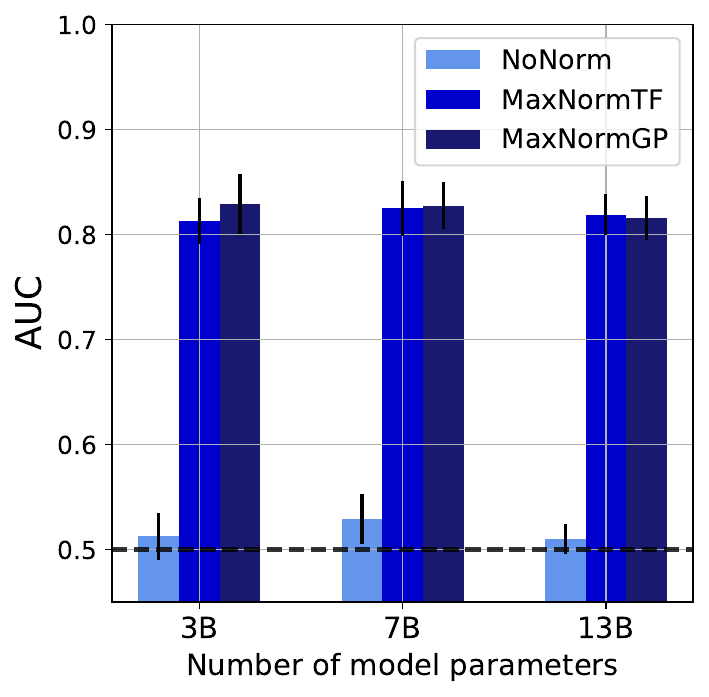} 
\includegraphics[width=0.49\linewidth]
{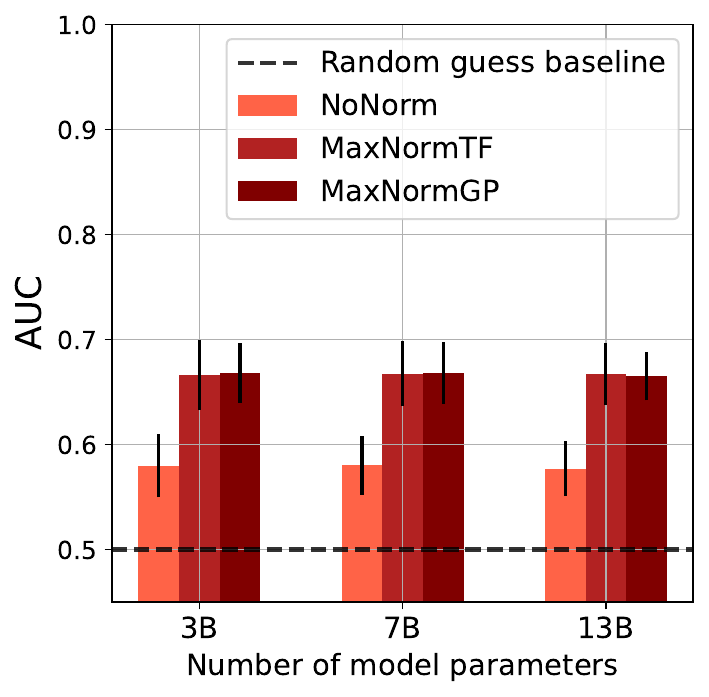} 
    \caption{Mean AUC for $C=2048$ across model sizes, for books from Project Gutenberg (left) and ArXiv papers (right).} 
\label{fig:across_models}
\end{figure}

Fig.~\ref{fig:across_models} shows that the AUC for membership classification remains highly consistent for even the smallest model. This implies that the memorization, as measured by our meta-classifier, for a "smaller" model of 3 billion parameters remains fairly similar. While these findings contrast with prior work~\cite{carlini2022quantifying,carlini2021extracting}, which show that memorization increases with model size, we note that our setup is significantly different. First, we only compare across larger models (3B+ parameters) and second, we measure memorization by evaluating document-level inference, which is very different than the extraction of specific sensitive information. Lastly, we also evaluate our setup using the model with 13 billion parameters and find that, again, the membership inference performance remains highly consistent. 

\subsection{Membership inference mitigations}
\label{sec:mitigations}
\textbf{Partial documents.} All results above consider the documents in their entirety, leveraging all token-level probabilities. In some cases, the entire document might not be known to the auditor, or there could be limitations in how many times the LLM can be queried. Hence, we here evaluate how the membership inference performance changes when only partial documents are considered. 

\begin{figure}
\centering
\includegraphics[width=0.49\linewidth]{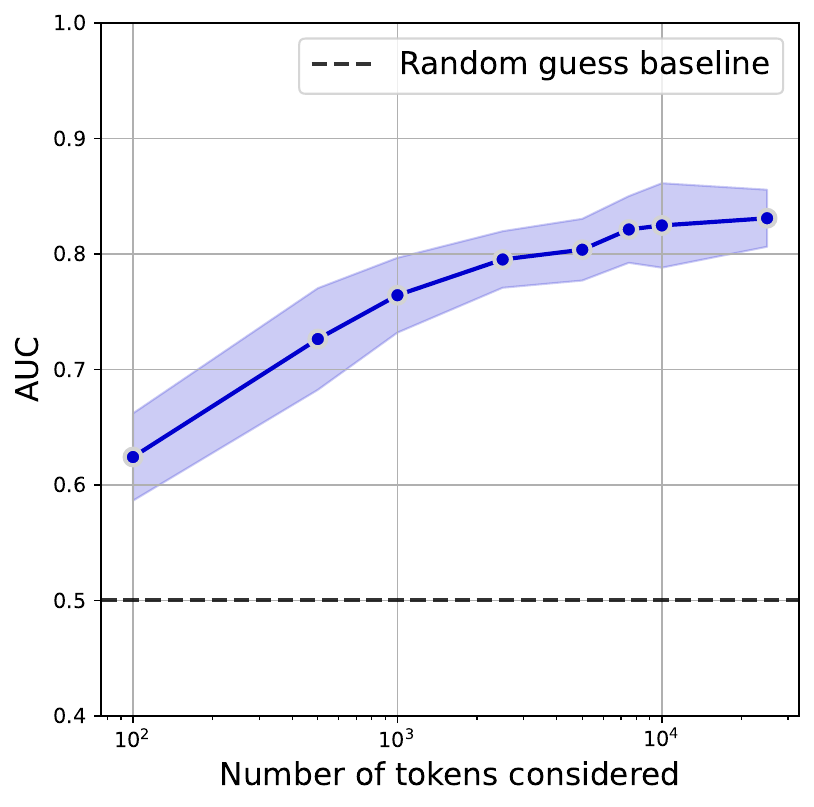} 
    \caption{Mean and standard deviation AUC for books from Project Gutenberg for an increasing number of tokens in a random excerpt sampled from the document.} 
\label{fig:partial_document}
\end{figure}

Fig.~\ref{fig:partial_document} shows how the AUC varies for an increasing number of tokens taken into account for books from Project Gutenberg using the best setup from Table~\ref{tab:books_primary}. In order to cleanly measure the impact of the number of tokens, we take a maximum number of tokens of $25,000$ and only consider books that have at least this number of tokens, eliminating $13.2\%$ of the books. For a given number of tokens, a random excerpt of this length is sampled from the book and then considered as the full document to run the membership inference. 

We find that even for $100$ tokens a mean AUC of $0.62$ is achieved, showing that even for smaller paragraphs sampled from documents our proposed method performs better than a random guess baseline. The AUC however steadily increases for an increasing number of tokens up until $0.83$ for $25,000$ tokens, effectively showcasing how considering a larger part of the document can significantly increase the membership inference performance. 

\textbf{Quantized models.} All results above consider the OpenLLaMA-7B in floating point precision of 16 bits. We here consider querying the language model with varying precision and evaluate the membership inference performance for books using the best setup from Table~\ref{tab:books_primary}. From the results in Table~\ref{tab:quantization}, we observe similarly good membership inference performance when querying the language model with lower precision. We hypothesize that the signal for membership picked up by our method likely does not change drastically with model precision when aggregating predicted probabilities over often more than $100,000$ tokens. 

\begin{table}[ht]
    \centering
    \begin{tabular}{cc}
    \toprule
        Precision & AUC \\
        \midrule 
         \textit{int4} & $0.841 \pm 0.025$ \\ 
        \cmidrule{1-2}
         \textit{int8} & $0.835 \pm 0.023$ \\ 
        \cmidrule{1-2}
         \textit{float16} & $0.856 \pm 0.021$ \\ 
        \cmidrule{1-2}
         \textit{float32} & $0.843 \pm 0.028$ \\ 
        \bottomrule
    \end{tabular}
    \caption{Mean and standard deviation AUC for the best setup (Tab~\ref{tab:books_primary}), querying OpenLLaMA-7B with varying precision.}
    \label{tab:quantization}
\end{table}

\subsection{Performance difference between datasets}

The results in Tables~\ref{tab:books_primary} and~\ref{tab:arxiv_primary} show a significant difference between the document-level membership inference performance on the books versus papers, with the highest mean AUC across setups of $0.856$ and $0.678$, respectively. We here discuss multiple hypotheses for this difference. 

\begin{figure}[ht]
\centering
\subfigure{
\includegraphics[width=0.45\linewidth]{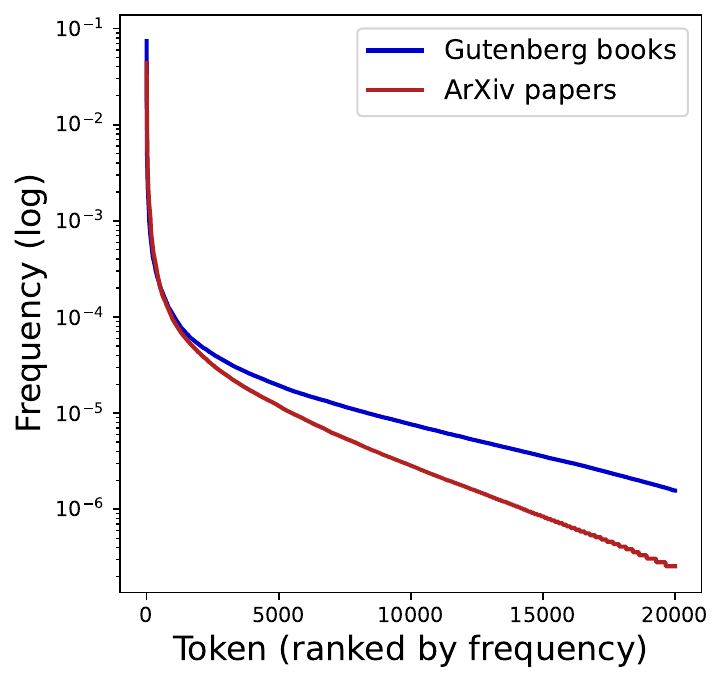} 
}
\label{fig:token_freq}
\subfigure{
\includegraphics[width=0.45\linewidth]{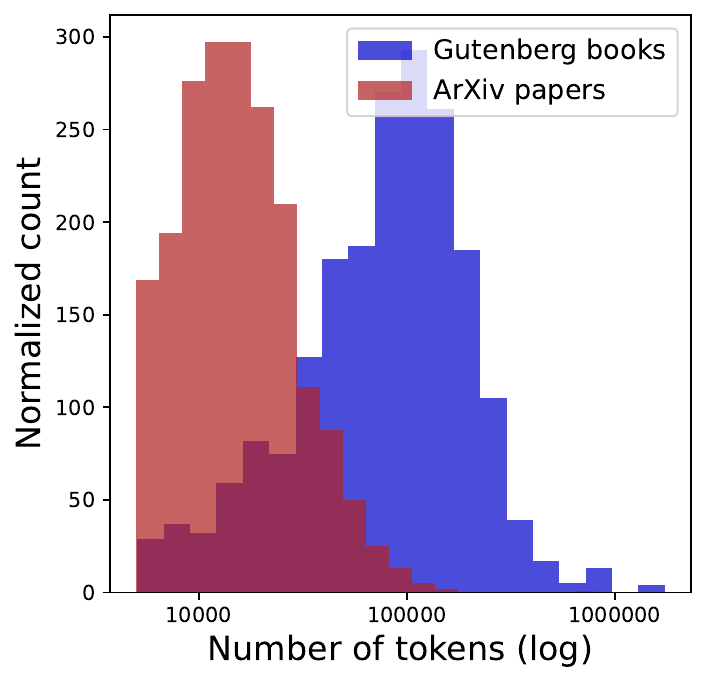}}
\caption{(a) Distribution of token frequency $R^{TF}$ for the top 20,000 tokens. (b) Density distribution of number of tokens per document. Both contain results for 2,000 documents.} 
\label{fig:papersVSbooks}
\end{figure}

First, the data itself is inherently different. The papers from ArXiv consist of data from raw LaTeX files. This contains a highly specific set of characters (e.g. LaTeX formatting, table content) while the natural language included in books from the literature is expected to contain a wider diversity of tokens. Fig.~\ref{fig:papersVSbooks} (a) compares how the token frequency $R^{TF}$ is distributed across the top 20,000 tokens between papers and books. This confirms our hypothesis that LaTeX papers contain more of a limited set of tokens, while the tokens used in books are more widely spread over the entire vocabulary $\vocabulary$, possibly affecting the distribution of predicted probabilities that our method requires as input.

Second, academic papers and books have a different length, i.e. total number of tokens per document. Fig.~\ref{fig:papersVSbooks} (b) confirms that the books considered in our experiments consist of significantly more tokens than the ArXiv papers, with an average of approximately 112,000 and 19,000 tokens respectively. 
This means that the language model has seen more tokens from member-books than from member-papers. Moreover, books typically contain repeated occurrences of characters, e.g. \emph{Harry Potter}, across the document. These characters would likely be rare tokens, occurring more often for the longer books than for papers, likely impacting the model memorization more for books than for papers and thus also the classification performance. Further, the document length could also play a role at inference time of the meta-classifier. Indeed, during feature extraction, more rare tokens contribute to our final feature set in the case of books than for papers, which again would impact the classification performance. 

Third, we anticipate that books in Project Gutenberg often represent the pieces of literature that are wide-spread across the internet. This means that there might exist an overlap between the books as included in the PG-19 dataset for training and the data scraped from the internet such as Common Crawl~\cite{commoncrawl} and C4~\cite{raffel2020exploring} which are also included in the training of OpenLLaMA~\cite{openlm2023openllama}. This leads to a potential level of duplication, which is reasonably more likely for books that have been added to Project Gutenberg earlier than later (i.e. members). This will likely impact the memorization of these books in $\languagemodel$ while, in contrast, academic papers, especially in LaTeX, are rarely distributed widely.


\section{Discussion}
We here construct a novel setup for document-level membership inference for real-world LLMs. 

First, we introduce a general paradigm to construct a labeled dataset for document-level membership that we believe to be realistic in practice. Indeed, LLMs typically use similar data sources to construct a training dataset, such as Common Crawl~\cite{commoncrawl}, C4~\cite{raffel2020exploring} and Project Gutenberg~\cite{projectgutenberg}. This makes the identification of member documents fairly feasible in practice. Moreover, these sources are typically continuously updated and timestamped, which enables an auditor to retrieve similar documents that were added after the LLM training dataset was created, i.e. a comparable set of non-members. 

This, however, depends on how recently a model was released and how frequently the data source of interest is updated. For ArXiv, more than $10,000$ new papers are added monthly~\footnote{\url{https://arxiv.org/stats/monthly_submissions}}. Hence, the retrieval of a sufficient amount of non-members (we here consider $1,000$) should be feasible shortly after the model release, especially as training data is typically collected some time prior to the model release. For Project Gutenberg, we find that a handful of new books are added daily~\footnote{\href{https://www.gutenberg.org/ebooks/search/?sort_order=release_date}{Project Gutenberg release date}}. While less frequent, a reasonable amount of non-members could still be collected not too long after the model release. 

Using this paradigm, we created a dataset of members and non-members for OpenLLaMA~\cite{openlm2023openllama} that we could verify using the publicly available RedPajama-Data~\cite{together2023redpajama}. We leave for future work how the meta-classifier performance changes when the full knowledge of training data is not available. In particular, our setup could be applied to the original LLaMA~\cite{touvron2023llama} model and potentially even on LLaMA-2~\cite{touvron2023llama}. 

In Sec.~\ref{sec:model-size-evaluation} we demonstrate the effectiveness of our method to language models of different model sizes (3B, 7B and 13B parameters) separately. In practice, it is however common that the entire model 'family' is trained on exactly the same dataset. We hypothesize that the auditor could construct a meta-classifier that takes as input document-level features queried from models of different sizes to make a potentially more accurate prediction for membership in practice.

In Sec.~\ref{sec:mitigations}, we find that when the target language model is queried with less precision, the membership inference remains accurate. We note, however, that we here exclusively consider pre-trained models, and leave for future work how alignment methods~\cite{zhang2023instruction,bai2022training} or watermarking~\cite{kirchenbauer2023watermark}, when applied to the language model, could alter these results. 

Further, we constrained ourselves to use a dataset of members $D_\text{M}$ and non-members $D_\text{NM}$ of size $1000$ each. With our computational resources, it takes us approximately a day to generate the predicted probabilities for each token in the books dataset by querying language model $\languagemodel$. We expect that researchers with a large resource pool could explore the impact of having larger datasets of members and non-members on the performance of the meta-classifier $\metaclassifier$.

Lastly, with this proof-of-concept, we hope to provide a way to retrieve a reasonable estimate of whether a document has been included in the training dataset of an LLM. Not only does this improve our understanding of memorization in ever larger models in practice, it also encourages increased transparency in what LLMs are trained on.


\section{Related work}
There exists a significant literature focusing on privacy attacks against (large) language models, which are relevant to our setup for document-level membership inference. 

Carlini et al.~\cite{carlini2019secret} proposes a framework to quantitatively asses unintended memorization in generative sequence models, such as language models trained for next token prediction. They effectively showcase how log-perplexity from a LSTM recurrent neural network trained for next token prediction can be used to extract potentially sensitive sequences from the training data (such as emails or credit card information). 
Further, Song et al.~\cite{song2019auditing} proposes a method to audit text generation models. Specifically, they use the shadow modeling technique to train a binary classifier for membership of the training data. For text generation, they implement this on the user level, aggregating predictions from individual pieces of text associated with the same user. 
While conceptually similar to the document-level membership inference in this paper, Song et al. apply their approach on recurrent neural networks trained on small datasets. Due to the computational cost to train a state-of-the-art, real-world large language model~\cite{radford2019language,brown2020language,touvron2023llama,touvron2023llama2,jiang2023mistral}, the use of the shadow modeling technique is no longer feasible. 
Further, the size of the dataset used to train these models~\cite{bender2021dangers} has drastically increased, making the contribution of an individual document distinctly harder to distinguish. For both reasons, new methods such as we here propose, are required to infer document-level membership against the most recent models.

Subsequently to their prior work, Carlini et al.~\cite{carlini2021extracting} introduces a novel method for training data extraction from the transformer-based GPT-2, with up to 1.5 billion parameters trained on a vast training dataset of 40GB of text~\cite{radford2019language}. Importantly for this paper, they show that a combination of perplexity queried from the model and zlib entropy (which gives a certain reference notion of surprise for a certain sequence of characters) allows for effective extraction of training data. Additionally, they show that larger models, i.e. neural networks with more parameters, tend to memorize more from their training data.
Mattern et al.~\cite{mattern2023membership} proposes a neighborhood attack for MIAs against language models. They also attack GPT-2, although now fine-tuned on specific data for which membership is then inferred. They use the difference between the target model loss computed on a given sequence and the loss computed on neighboring samples, obtained by replacing tokens using masked language models.

In contrast with the prior contributions mentioned above, this paper focuses specifically on document-level membership inference and applies it to real-world, large language models (7B+ parameters). This is, to the best of our knowledge, the first work of its kind.

Further, researchers have proposed defenses against these privacy attacks, such as de-duplication of the training data~\cite{kandpal2022deduplicating} or differentially private training~\cite{li2021large}. However, Lukas et al.~\cite{lukas2023analyzing} challenges this, stating that these defenses do not reduce the risk posed to personal identifiable information.
Lastly, various other works have developed membership inference attacks on more domain as well as task-specific language models, such as clinical language models~\cite{jagannatha2021membership} or language models in non-English languages~\cite{oh2023membership}, or fine-tuned models for machine translation~\cite{hisamoto2020membership} or classification~\cite{shejwalkar2021membership,shachor2023improved}.

\label{sec:related_work}

\section{Conclusion}

This paper proposes the first setup and methodology for document-level membership inference for real-world LLMs. 

First, we formalize the task and introduce a procedure to realistically construct a labeled training dataset. For non-members, we rely on original documents that are similar to documents seen by the model but made available after the release date of the model. We then construct a dataset for the OpenLLaMA~\cite{openlm2023openllama} model for both books from Project Gutenberg~\cite{projectgutenberg} and academic papers from ArXiv.

We then propose a methodology to infer document-level membership. We find that the distribution of predicted probabilities for all the tokens within a document, normalized by a value reflecting the rarity of the token, contains meaningful information for membership. Indeed, in the best performing setup, the meta-classifier infers binary membership with an AUC of 0.86 and 0.68 for books and papers respectively. This suggests our classifier's ability to accurately infer whether a document has been seen by the LLM during training. 


\section*{Availability}
The code used to generate the results in this paper has been made publicly available on Github\footnote{\url{https://github.com/computationalprivacy/document-level-membership-inference}}.

\bibliographystyle{plain}
\bibliography{bibilography}

\end{document}